\pdfoutput=1

\documentclass[11pt]{article}

\usepackage[preprint]{acl}

\usepackage{times}
\usepackage{latexsym}
\usepackage{textcomp}
\usepackage[T1]{fontenc}

\usepackage[utf8]{inputenc}

\usepackage{microtype}

\usepackage{inconsolata}

\usepackage{graphicx}

\usepackage{booktabs}
\usepackage{float}
\usepackage{xcolor}
\usepackage{colortbl}

\usepackage{multicol}
\usepackage{multirow}
\usepackage{makecell}
\usepackage{tabularx}
\usepackage{subcaption}

\usepackage{listings}
\definecolor{our_gray}{HTML}{E5E5E5}
\definecolor{our_red}{HTML}{F8D5D8}
\definecolor{our_green}{HTML}{E4F7C1}
\definecolor{our_blue}{HTML}{C2F3FF}

\usepackage{tcolorbox}

\definecolor{commentgray}{RGB}{130, 130, 130}
\definecolor{stringblue}{RGB}{32, 74, 135}
\definecolor{keywordpurple}{RGB}{106, 13, 173}
\definecolor{classname}{RGB}{173, 23, 23}
\definecolor{entitycolor}{RGB}{150, 50, 50}
\definecolor{bordergray}{RGB}{200, 200, 200}

\lstdefinestyle{gollie}{
    language=Python,
    basicstyle=\ttfamily\footnotesize,
    keywordstyle=\color{keywordpurple}\bfseries,
    commentstyle=\color{commentgray},
    stringstyle=\color{stringblue},
    emph={BeBorn, LifeEvent, Entity, List}, emphstyle=\color{classname}, emphstyle=\color{classname},
    morekeywords={dataclass, List},
    frame=single,
    framesep=5pt,
    rulecolor=\color{bordergray},
    showstringspaces=false,
    breaklines=true,
    breakindent=0pt,
    breakatwhitespace=false
}
\newcommand{\highlight}[2]{%
    \begingroup
    \definecolor{hlcolor}{HTML}{#1}%
    \sethlcolor{hlcolor}%
    \hl{#2}%
    \endgroup
}

\definecolor{classgreen}{RGB}{34,139,34}
\definecolor{bluekeyword}{RGB}{0,0,255}
\definecolor{purpletype}{RGB}{106,13,173}
\definecolor{blackbold}{RGB}{0,0,0}

\lstdefinestyle{custompython}{
    language=Python,
    basicstyle=\ttfamily\footnotesize,  
    backgroundcolor=\color{gray!5}, 
    frame=single, 
    rulecolor=\color{gray}, 
    keywordstyle=[1]\color{blue}\bfseries, 
    keywordstyle=[2]\color{cyan}\bfseries, 
    keywordstyle=[3]\color{violet}\bfseries, 
    emph={BeBorn, LifeEvent, Extradite, JusticeEvent}, emphstyle=\color{magenta}, 
    morekeywords={[1]class, def, return, if, else, elif, for, while, try, except, import, from, as, with},
    morekeywords={[2]@dataclass}, 
    morekeywords={[3]str, List, Dict, Tuple, Set}, 
    stringstyle=\color{violet}, 
    commentstyle=\color{darkgray}, 
    numberstyle=\tiny\color{gray}, 
    stepnumber=1, 
    showstringspaces=false,
    breaklines=true, 
    captionpos=b 
}

\usepackage{listings} 
\usepackage{graphicx} 

\lstdefinelanguage{json}{
  basicstyle=\ttfamily\small,
  showstringspaces=false,
  breaklines=true,
  literate=
    *{0}{{{\color{orange}0}}}{1}
     {1}{{{\color{orange}1}}}{1}
     {2}{{{\color{orange}2}}}{1}
     {3}{{{\color{orange}3}}}{1}
     {4}{{{\color{orange}4}}}{1}
     {5}{{{\color{orange}5}}}{1}
     {6}{{{\color{orange}6}}}{1}
     {7}{{{\color{orange}7}}}{1}
     {8}{{{\color{orange}8}}}{1}
     {9}{{{\color{orange}9}}}{1}
     {:}{{{\color{orange}{:}}}}{1}
     {,}{{{\color{orange}{,}}}}{1}
}

\lstdefinestyle{customjson}{
  language=JSON,
  basicstyle=\ttfamily\small,
  backgroundcolor=\color{gray!5}, 
  keywordstyle=\color{black}, 
  stringstyle=\color{violet}, 
  commentstyle=\color{cyan}, 
  morecomment=[l][\color{magenta}]{//}, 
  numberstyle=\tiny\color{gray}, 
  stepnumber=1, 
  breaklines=true, 
  frame=single, 
  rulecolor=\color{gray}, 
  captionpos=b 
}

\usepackage{soul}
\usepackage{textcomp}
\usepackage{pifont}
\usepackage{stfloats} 
\usepackage{float}
\usepackage{amssymb}

\definecolor{darkgreen}{rgb}{0.0, 0.5, 0.0}

\title{Instruction-Tuning LLMs for Event Extraction with Annotation Guidelines}

\author{
  Saurabh Srivastava$^*$, 
  Sweta Pati$^*$, 
  \textbf{Ziyu Yao}\\
  George Mason University, Fairfax, VA \\
  \{ssrivas6, spati, ziyuyao\}@gmu.edu
}

\begin{document}
\maketitle

\def\thefootnote{*}\footnotetext{The first two authors contribute equally.}\def\thefootnote{\arabic{footnote}}

\begin{abstract}




In this work, we study the effect of annotation guidelines---textual descriptions of event types and arguments, when instruction-tuning large language models for event extraction. We conducted a series of experiments with both human-provided and machine-generated guidelines in both full- and low-data settings. Our results demonstrate the promise of annotation guidelines when there is a decent amount of training data and highlight its effectiveness in improving cross-schema generalization and low-frequency event-type performance.\footnote{Our source code and datasets are available at \href{https://github.com/Ziyu-Yao-NLP-Lab/PyCode-TextEE}{https://github.com/Ziyu-Yao-NLP-Lab/PyCode-TextEE}.}

\end{abstract}

\section{Introduction}


Event Extraction (EE) aims to identify and structure \textit{what, who, when, where, and how} of real-world events from given textual resources~\cite{doddington-etal-2004-automatic, ji-grishman-2008-refining, li2022survey, xu2024large}. 
Translating this abstraction requires complex schema specifications that define event types, argument roles, and their interrelationships, yet being able to precisely capture the language nuances and distinguish between event types and argument roles, which posits the task as an inherently challenging problem.
Recently, large language models (LLMs) have transformed NLP research and practices dramatically, owing to the rich knowledge and other capabilities (e.g., reasoning) they have obtained from extensive pre-training~\cite{wei2022chain, chen2023program, shi2023dont}. This transformation has similarly impacted the broader research field of Information Extraction (IE). Existing applications of LLMs to IE can be categorized into two lines. The \emph{prompt engineering-based approaches}, often based on proprietary LLMs, consider an LLM as a black box, querying it with task specifications via zero- or few-shot prompting and relying on its latent knowledge to extract interested information~\cite{gao2023exploringfeasibilitychatgptevent, wang2023code4struct, li2023evaluating, srivastava-etal-2023-mailex}. However, these approaches not only lead to inferior performance but also incur prohibitive costs, especially when the task is complex.

Our work will thus focus on the second line of approach, namely, \emph{instruction-tuning open-weight LLMs}. This line of approach adapts an LLM to specific IE tasks and schemas by directly training it to follow the task instructions, which offers a promising yet cost-effective solution. For example, \citet{wang2023instructuiemultitaskinstructiontuning}  leverages natural language instructions to
guide large language models for IE tasks; \citet{li-etal-2024-knowcoder} proposed a two-phase learning framework that enhances schema understanding and following ability via automatically annotated data. More recently, \citet{sainz2024gollie} instruction-tuned LLaMA~\cite{touvron2023llamaopenefficientfoundation} on multiple IE datasets and discovered \emph{annotation guidelines}---textual descriptions of an event type and its argument roles used by human annotators when collecting the dataset, as effective components of an IE task's instruction. Despite the promise of the existing explorations, however, most of them have focused on the relatively simpler task of Named Entity Recognition, yet how to properly instruction-tune LLMs for the structured EE task is still understudied.
\begin{figure*}[t!]
    \centering
    \includegraphics[width=.95\linewidth]{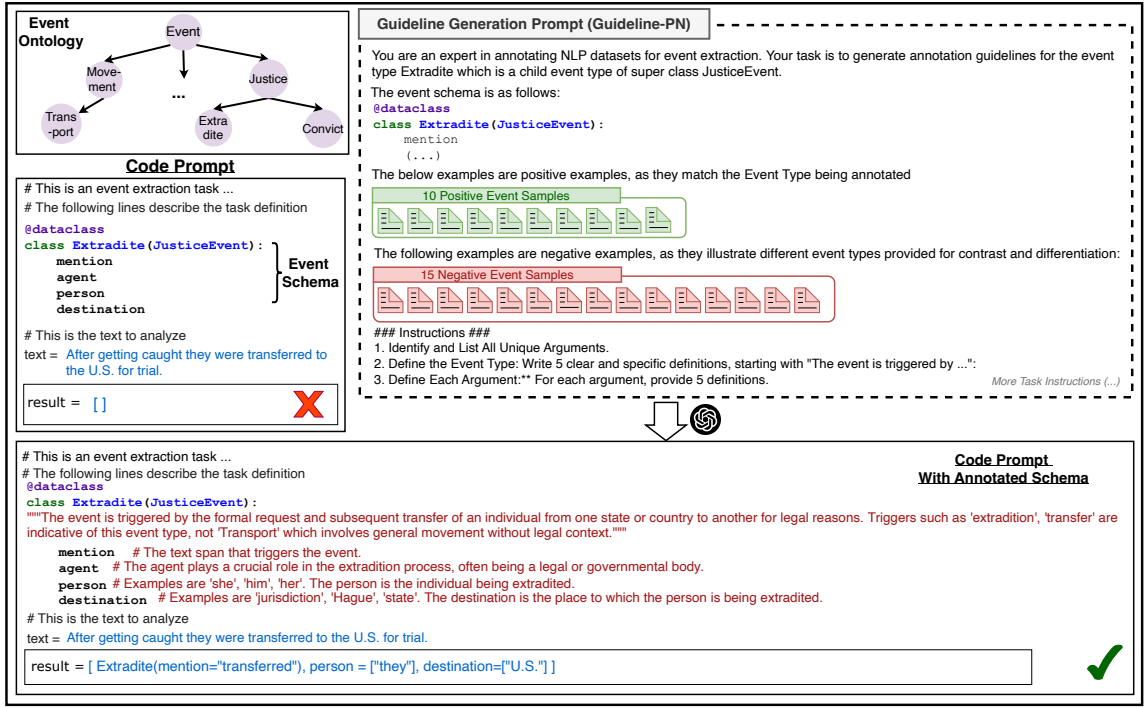}
    \caption{Overview of our exploration of automatically generating annotation guidelines to augment code-format instruction tuning for EE. Prompt template for Guideline-PN and the example outputs are shown.
    }
    \label{fig:overview}
\end{figure*}

To fill this gap, we study instruction-tuning LLMs for EE, with a focus on the role of annotation guidelines in task instructions (Fig. \ref{fig:overview}). 
We conduct a systematic analysis using LLaMA-3.1-8B-Instruct on two EE datasets (ACE05~\cite{doddington-etal-2004-automatic} and RichERE~\cite{song-etal-2015-light}) under varied training settings. Our key findings are organized around four themes:

1) \textit{Effect of Annotation Guidelines on Event Extraction} — We found that annotation guidelines improve performance by helping the model distinguish fine-grained event types. However, this advantage may diminish when negative sampling is introduced during training, which allows the model to learn event distinctions from additional contrastive examples instead.

2) \textit{Comparing Machine-Generated and Human-Written Guidelines} — Prior work assumed access to human-authored guidelines, which may not hold in practice. We thus proposed 5 different ways to automatically generate annotation guidelines. We find that they outperform human-written ones by up to 11\% and 7\% in trigger and argument classifications, respectively. 

3) \textit{Guidelines in Data-Scarce Scenarios} — Our results show that with only 2000 training samples, guidelines allow LLMs to reach performance levels comparable to full-data training. However, when data is extremely scarce (100 samples), models tend to rely more on memorization than on guideline-driven schema constraints.

4) \textit{Cross-Schema Generalization} — We assess whether structured guidelines help models generalize to different EE schemas. While models trained on RichERE transfer well to ACE (suggesting fine-to-coarse schema adaptation is feasible), the reverse scenario sees a performance drop due to RichERE’s more complex event structures and expanded argument roles.

{Finally, we validated the consistency of our findings across both model scale and diversity. Using the smaller LLaMA-3.2-1B-Instruct model, we observed that annotation guidelines retain their benefits, reducing common EE errors and supporting both frequent and rare event types. Extending beyond the LLaMA family, experiments with Qwen2.5-Coder-1.5B-Instruct \cite{hui2024qwen25codertechnicalreport}, a code-oriented model with a distinct pertaining objective, revealed similarly robust gains. We also evaluated on Speed++ \cite{parekh-etal-2024-speed}, a domain-shifted dataset of informal epidemic-related tweets, where guidelines continued to significantly improve performance. Together, these results confirm the broad utility of machine-generated guidelines across model architectures, data regimes, and textual domains.}

\section{Approach}\label{sec:approach}


\subsection{Task Formulation}
\label{sec:formulation}
Given an input sentence \( X \), the goal of EE is to extract the structured event information $Y$ from the sentence, adhering to predefined schema constraints $\mathcal{E}$. The extraction task consists of (1) \textbf{Trigger Extraction}, which localizes an event trigger span and classifies its event type, and (2) \textbf{Argument Extraction}, where the task is to identify spans in \( X \) that serve as argument roles within the extracted event instance.  

When an autoregressive LLM is tasked with EE, the extraction of event instances is formulated in a generative way, with the LLM generating a sentence describing the extracted event instances. Specifically, the prompt to the LLM is defined as  $P = [I \oplus {E}_e \oplus X]$, where \(\oplus\) is the concatenation operation, \( I \) represents the {task instruction}, which specifies the structured output format and task definition, and \( {E}_e \in \mathcal{E}\) denotes the {event schema} of an interested type $e$ from a predefined set $\mathcal{E}$.

Let \( \mathcal{D} = \{(e_i, X_i, Y_i)\}_{i=1}^{N} \) denote a dataset of annotated event examples, where each \( X_i \) corresponds to a prompt instance \( P_i \) for the interested event type $e_i$. The objective function of instruction tuning for EE is as follows: $\mathcal{L}(\mathcal{D}; \theta) = - \sum_{i} \sum_{j} \log p_{\theta} (Y_{ij} \mid P_i, Y_{i,<j})$,
where \( Y_{i,<j} \) represents previously generated tokens in the structured output sequence, ensuring an autoregressive formulation. 

Existing work identified the structure of EE outputs to be critical~\cite{jiao-etal-2023-instruct, wang2023code4struct}. In particular, \citet{wang2023code4struct} found that formulating the EE output in a \emph{code format} can {take advantage of Programming Language features such as inheritance and type annotation to introduce external knowledge or add constraints.} In our work, we follow the same formatting strategy and represent the EE task as a code generation problem. Specifically, the event schema ${E}_e$ is represented as a Python class; accordingly, every extracted event instance is represented as a Python object of the corresponding event class. When there are multiple event instances implied in the input $X$, a list of Python objects will be generated; when there is no event specified in $X$, we expect an empty Python list to be the model output. An example is shown in Figure~\ref{fig:overview}.

During training, we provide only the ground-truth event schema in the prompt; when the text input $X$ does not include any event, a random event schema will be chosen. 
At inference time, given a test instance $X$, we pair the input with every possible event type in the schema set $\mathcal{E}$, prompt the LLM to extract any implied event instances, and perform model evaluation based on the aggregated extraction outputs. As such, a well-performing LLM needs both extract the complete event instances and avoid events that are not indicated in $X$.

\subsection{Instruction-Tuning LLMs with Annotation Guidelines}
\label{sec:guideline_generation}

Recent work by \citet{sainz2024gollie} demonstrated the effectiveness of integrating annotation guidelines in the code-format instructions of IE tasks. Specifically, when describing the event type schema $E_e$, a textual description is added to the event type and each of its argument roles (Figure~\ref{fig:overview}). As such, the LLM is expected to more easily understand the meaning of the event type while being instructed to extract any occurring events from the input $X$. While \citet{sainz2024gollie} evaluated annotation guidelines in the broad IE task, their main focus has been on Named Entity Recognition, rather than the complicated EE task. 
Furthermore, their approach assumed the availability of pre-existing human-curated guidelines, an assumption that may not always hold in real-world applications. To bridge this gap, we explore methods to automatically generate annotation guidelines and assess their effectiveness in comparison to human-authored ones.\footnote{Human-written guidelines for ACE05 are available \href{https://www.ldc.upenn.edu/sites/www.ldc.upenn.edu/files/english-events-guidelines-v5.4.3.pdf}{here}.}
\begin{table*}[t!]
\centering
\resizebox{.95\linewidth}{!}{%
\begin{tabular}{>{\centering\arraybackslash}p{3.8cm}p{17cm}}
    \toprule
    \multicolumn{2}{c}{\textbf{Examples of Annotation Guidelines for Event Type: Extradite (ACE05)}} \\
    \midrule
    \textbf{\textsc{Guideline-H}}\newline \small{Avg. Length - 107.67 tokens} & 
    {\textbf{Event Type:} {{An EXTRADITE Event occurs whenever a PERSON is sent by a state actor from one PLACE to another place for the purposes of legal proceedings there. }}\newline
    \textbf{Arguments:} \newline 
    - \textsc{Agent}: The extraditing agent.  \newline 
    - \textsc{Person}: The person being extradited.}\\
    \midrule
    \textbf{\textsc{Guideline-P}}  \newline \small{Avg. Length - 163.87 tokens}&
    {\textbf{Event Type:} The Extradition event refers to the formal process where one jurisdiction delivers a person accused \textbf{(...)} \highlight{D5E8D4}{The event can be triggered by terms such as `extradition' \textbf{(...)}} \highlight{BCD4E6}{Edge cases include situations where the term `extradition' is used metaphorically or in a non-legal context.} \newline
    \textbf{Arguments:} \newline 
    - \textsc{Agent}:\textbf{(...)} the agent is the organization or authority \textbf{(...)}. \highlight{D5E8D4}{Examples include `court', `government', \textbf{(...)}} \newline 
    - \textsc{Person}: \textbf{(...)} individual who is being transferred to another jurisdiction. \highlight{D5E8D4}{Examples are `she', \textbf{(...)}}
    }\\
    \midrule
    \textbf{\textsc{Guideline-PN}} \newline \small{Avg. Length - 285.24 tokens}&
    \textbf{Event Type:} The event is triggered by the formal request \textbf{(...)} for legal reasons. \highlight{D5E8D4}{Triggers such as `extradition' are indicative of this event type}, \highlight{FACEC6}{not `Transport' which involves general movement without legal context.} \\
    & \textbf{Arguments:} \\
    & - \textsc{Agent}: The agent is responsible for the legal and procedural aspects of the extradition,\textbf{(...)}. \highlight{D5E8D4}{An example is `the original court' \textbf{(...)}} \\
    & - \textsc{Person}: \textbf{(...)} one who is being moved from one place to another under legal authority. \highlight{D5E8D4}{For example, `he' \textbf{(...)}} \\
    \midrule
    \textbf{\textsc{Guideline-PS}} \newline \small{Avg. Length - 159.79 tokens}&
    \textbf{Event Type:} \textbf{(...)} person being moved to a new jurisdiction \textbf{(...)}. \highlight{FACEC6}{This differs from events like `TrialHearing' or `Convict', which focus on the legal proceedings and outcomes within a single jurisdiction.} \\
    & \textbf{Arguments:} \\
    & - \textsc{Agent}: \textbf{(...)} \highlight{BCD4E6}{Edge cases may include international organizations or coalitions} \textbf{(...)} \highlight{D5E8D4}{such as the U.N. \textbf{(...)}} \\
    & - \textsc{Person}: \highlight{FACEC6}{Unlike the `defendant' in events like `TrialHearing' or `Convict', the person in the `Extradite' event is specifically being transferred for legal proceedings or punishment.} \\
    \midrule
    \textbf{\textsc{Guideline-PN-Int}} \newline \small{Avg. Length - 439.94 tokens}&
    \textbf{(...)} \highlight{D5E8D4}{Key triggers include terms like `extradite', `extradition', and `extraditing'.} \highlight{FACEC6}{It is distinct from events like `ArrestJail' and `ReleaseParole', as it specifically involves \textbf{(...)}} \\
    & \textbf{Arguments:} \\
    & - \textsc{Agent}: The agent \textbf{(...)} typically a legal or governmental body. \highlight{D5E8D4}{Examples include `court', `government'\textbf{(...)}} \\
    & - \textsc{Person}: The person is the individual being extradited, the subject of the legal transfer. \highlight{D5E8D4}{Examples include `she', `him', and `her'.} \\
    \midrule
    \textbf{\textsc{Guideline-PS-Int}} \newline \small{Avg. Length - 434.64 tokens}&
    The 'Extradite' event involves the legal transfer of a person \textbf{(...)}. \highlight{FACEC6}{It is distinct from events like `ArrestJail', \textbf{(...)}, and `ReleaseParole' or `Pardon', \textbf{(...)}} \\
    & \textbf{Arguments:} \\
    & - \textsc{Agent}: The agent is the entity \textbf{(...)} \highlight{D5E8D4}{such as a court, government, or police department.} \highlight{BCD4E6}{This entity ensures the transfer is conducted according to legal protocols \textbf{(...)}} \\
    & - \textsc{Person}: \textbf{(...)} They are the central figure in the extradition process, \highlight{FACEC6}{distinct from a `defendant' in other legal events, \textbf{(...)}}\highlight{BCD4E6}{ This may include high-profile individuals or groups.} \\

    \bottomrule
    \end{tabular}
    }
    \caption{Examples of annotation guidelines for the event type \texttt{Extradite} from ACE05. Due to space limits, only \texttt{agent} and \texttt{person} were shown for arguments, and only 1 out of the 5 guideline samples were shown for \textbf{P}, \textbf{PN}, and \textbf{PS}.
    We highlight \highlight{FACEC6}{distinctions from other event types}, \highlight{D5E8D4}{example mentions}, and \highlight{BCD4E6}{edge cases} in guidelines.
    }
    \label{tab:guideline-examples}
\end{table*}

To develop a scalable and cost-effective approach for guideline generation, we employ a reverse engineering strategy, leveraging both annotated event examples and the strong generative capabilities of LLMs. As illustrated in Figure \ref{fig:overview}, we construct a guideline generation prompt for each event type $e$ by providing a few annotated examples $\{(X_i, Y_i)\}$ demonstrating the existence or non-existence of event instance of type $e$, and then prompt an LLM (GPT-4o in our experiment) to generate annotation guidelines for $e$. In total, we experimented with five variants of machine-generated guidelines:
\textbf{(1) Guideline-P:} We prompt the LLM with 10 positive annotated examples of type $e$ to generate the annotation guidelines. Inspired by \citet{sainz2024gollie}, we sample 5 distinct guidelines for each event type, which can be used during the model training to ensure that the model is exposed to multiple rephrasings of the guidelines rather than memorizing and overfitting to a specific one.
\textbf{(2) Guideline-PN (Positive + Negative Examples):} In addition to 10 positive event annotations, we also provide 15 negative annotations where the input $X$ does not imply event instances of type $e$. Similarly, we prompt the LLM to generate 5 distinct guidelines for each event type.
\textbf{(3) Guideline-PS (Positive + Sibling Events):} Similar to Guideline-PN, we prompt the LLM with both positive and negative event annotations. However, the negative annotations are selected from the sibling event types of the target type $e$ (e.g., Arrest vs. Jail), as defined by the event ontology. We hypothesize that the critical challenge for EE lies in distinguishing between sibling event types; hence, an instructed LLM can benefit from following annotation guidelines that particularly emphasize the difference between sibling event types. As in the earlier variants, we generate 5 guideline samples per event type.
\textbf{(4) Guideline-PN-Int} and \textbf{(5) Guideline-PS-Int:} Finally, we create two more variants that \underline{Int}egrate the 5 diverse guideline samples from Guideline-PN and Guideline-PS into a comprehensive one, respectively. 
Examples of the 5 guideline variants are shown in Table~\ref{tab:guideline-examples}.
The prompt templates used for generating guidelines and example generations
are provided in Appendix\hyperref[sec:prompt-design]{~C} and~\ref{sec:app_dd}, respectively.

\section{Experiments}

\subsection{Experimental Setup} 
\paragraph{Datasets.}
We perform experiments on two standard EE datasets: \textbf{ACE05} \cite{doddington-etal-2004-automatic} and \textbf{RichERE} \cite{song-etal-2015-light}. Both of them exhibit fine-grained event distinctions, and RichERE includes sparser event annotations (i.e., fewer event-labeled sentences), which makes it more challenging. Moreover, RichERE does not come with human-written annotation guidelines. Datasets were split following the TextEE benchmark~\cite{huang2024textee} and then converted to code format automatically by our scripts. 


\paragraph{Evaluation.} 
Following prior work~\cite{huang2024textee}, we evaluate the model on four F1 metrics:
\textbf{(1) Trigger Identification (TI)}, which measures correct trigger span extraction, \textbf{(2) Trigger Classification (TC)}, which additionally requires event-type correctness, \textbf{(3) Argument Identification (AI)}, which ensures correct argument role association with the predicted trigger, and \textbf{(4) Argument Classification (AC)}, which further requires role-type correctness and is thus the most comprehensive metric on a model's EE performance. When evaluating the model on the Guideline-P, PN, and PS variants, one guideline is randomly selected each time.

As a side benefit of representing events in a structured code format, we can easily evaluate an extracted event instance by directly instantiating its corresponding Python object based on the event schema's Python class definitions, checking if the object is valid (e.g., missing arguments or including hallucinated arguments) and comparing it with the ground truth. This code-based evaluation thus prevents the tedious string-matching process adopted in prior work~\cite{li-etal-2021-document}. 

\paragraph{Model Training.} We experimented with the {LLaMA-3.1-8B-Instruct} model~\cite{grattafiori2024llama3herdmodels}, selected for its demonstrated proficiency in processing structured code-based inputs and generating coherent outputs. When instruction-tuning the model under the Guideline-P, PN, and PS variants, we randomly sample one of the generated guidelines, a strategy found to prevent the model from memorizing specific guidelines in ~\citet{sainz2024gollie}.
For parameter-efficient training, we implemented rsLoRA~\cite{kalajdzievski2023rankstabilizationscalingfactor} using the Unsloth framework~\cite{unsloth}. 

We include all details about datasets, evaluation, and model training in Appendix~\ref{sec:appendix_preprocessing}-\ref{sec:app_additional_details}. {We will open-source our scripts for automatically converting datasets in TextEE~\cite{huang2024textee} into Python code format (we dub the processed version as \textbf{PyCode-TextEE}) and for evaluating extracted events automatically via code execution at our GitHub repository \url{https://github.com/Ziyu-Yao-NLP-Lab/PyCode-TextEE}.}

\begin{table*}[th!]
\centering
\setlength{\tabcolsep}{4pt}
\renewcommand{\arraystretch}{1.2}
\definecolor{rowgray}{gray}{0.97} 
\definecolor{rowgray2}{gray}{1.0} 
\definecolor{lightblue}{RGB}{173, 216, 250}
\resizebox{\textwidth}{!}{%
\begin{tabular}{lcccc|cccc|cccc|cccc}
\toprule

\multirow{2}{*}{\textbf{Experiments}} & \multicolumn{4}{c|}{\textbf{ACE w/o NS}} & \multicolumn{4}{c|}{\textbf{ACE w/ NS}} & \multicolumn{4}{c|}{\textbf{RichERE w/o NS}} & \multicolumn{4}{c}{\textbf{RichERE w/ NS}} \\

\cmidrule(lr){2-5} \cmidrule(lr){6-9} \cmidrule(lr){10-13} \cmidrule(lr){14-17}
& \textbf{TI} & \textbf{TC} & \textbf{AI} & \textbf{AC} & \textbf{TI} & \textbf{TC} & \textbf{AI} & \textbf{AC} & \textbf{TI} & \textbf{TC} & \textbf{AI} & \textbf{AC} & \textbf{TI} & \textbf{TC} & \textbf{AI} & \textbf{AC} \\
\midrule
\rowcolor{lightblue}\textbf{NoGuideline}      & 39.57  & 39.57  & 31.05  & 29.73  & \textbf{84.15}  & \textbf{84.15}  & \textbf{64.99}  & \textbf{61.96}  & \underline{35.11}  & \underline{35.11}  & 27.16  & 25.32  & 42.27  & 42.27  & 32.38  & 31.56  \\
\rowcolor{rowgray2}\textbf{Guideline-H}      & 40.71  & 40.71  & 30.76  & 28.64  & 56.30  & 56.30  & 44.82  & 43.13 & --  & --  & --  & --  & --  & --  & --  & -- \\
\rowcolor{rowgray}\textbf{Guideline-P}      & \textbf{51.46}  & \textbf{51.46}  & \textbf{37.82}  & \textbf{35.20}  & 72.86  & 72.86  & 55.01  & 53.73  & 34.38  & 34.38  & \underline{28.04}  & \underline{26.35}  & 67.92  & 67.92  & 52.29  & 44.93 \\
\rowcolor{rowgray2}\textbf{Guideline-PN}     & \underline{49.60}  & \underline{49.60}  & 35.80  & 32.81  & \underline{80.77}  & \underline{80.77}  & \underline{63.20}  & \underline{60.34}  & \textbf{40.89}  & \textbf{40.89}  & \textbf{30.04}  & \textbf{27.18}  & \underline{75.35}  & \underline{75.35}  & \textbf{60.85}  & \textbf{57.10}  \\
\rowcolor{rowgray}\textbf{Guideline-PS}     & 47.93  & 47.93  & \underline{37.19}  & \underline{34.88}  & 79.23  & 79.23  & 59.00  & 56.88  & 32.41  & 32.41  & 24.63  & 22.78  & \textbf{76.45}  & \textbf{76.45}  & \underline{60.42}  & \underline{56.26}  \\
\rowcolor{rowgray2}\textbf{Guideline-PN-Int} & 40.17  & 40.17  & 30.46  & 28.34  & 51.95  & 51.95  & 41.09  & 39.32  & 27.11  & 27.11  & 21.93  & 20.81  & 42.40  & 42.40  & 33.22  & 31.67 \\
\rowcolor{rowgray}\textbf{Guideline-PS-Int} & 39.51  & 39.51  & 31.27  & 30.26  & 53.70  & 53.70  & 42.62  & 41.10  & 31.61  & 31.61  & 26.70  & 24.96  & 52.60  & 52.60  & 41.06  & 39.46  \\
\bottomrule
\end{tabular}%
}

\caption{Evaluation results (\%) for end-to-end EE tasks trained on complete train data. Models trained \textbf{with Negative Samples (w/ NS)} include negative example augmentation. (\textbf{Best} and \underline{Second Best} performances)}
\label{tab:full_train_table}
\end{table*}
\subsection{RQ1: Do the annotation guidelines allow an LLM to more precisely extract occurring events?}
To assess the impact of incorporating annotation guidelines in the EE instructions, we compare instruction-tuning an LLM with and without guidelines. We hypothesize that including the annotation guidelines can help the LLM more easily distinguish between similar event types. To understand its impact, we also compare this approach with a ``negative sampling (NS)'' approach. Specifically, we instruction-tune the LLM on an augmented training set, where each training example is supplemented with 15 randomly selected negative samples, i.e., triplets of $(e_{neg}, X, \phi)$ with non-existing event type $e_{neg}$ yielding empty extraction output. We note that annotation guidelines and negative sampling are two complementary approaches for an LLM to learn to distinguish between event types. In our experiments, we thus evaluated the effect of annotation guidelines in two independent settings: (1) training on the original training set (\textbf{w/o NS}) and (2) training on the negative sample-augmented training set (\textbf{w/ NS}). {Additional details on example selections are provided in Appendix \ref{sec:neg_details}.}

Table~\ref{tab:full_train_table} shows the results.
In the \textbf{w/o NS} setting, including annotation guidelines (\textbf{Guideline-P}, \textbf{PN}, and \textbf{PS}) consistently improves performance across both datasets. Our analysis in Section~\ref{sec:analyais} further validated that the guidelines indeed enable the LLM to understand the nuanced differences between event types. On {ACE w/o NS}, \textbf{Guideline-P} achieves the highest scores across all four metrics, leading to around 10\% TC and 5\% AC gains over \textbf{NoGuideline}.
Similarly, on {RichERE w/o NS}, \textbf{Guideline-PN} outperforms \textbf{NoGuideline} by about around 5\% TC and 2\% AC.  

Training the LLM with augmented negative samples, as we expected, helps the model better distinguish between event types; for example, \textbf{NoGuideline} in the \textbf{w/ NS} setting achieves 30\% higher AC on ACE and 6\% higher AC on RichERE, compared to its counterparts in the \textbf{w/o NS} setting. 
However, the effects of annotation guidelines in the \textbf{w/ NS} setting diverge between the two datasets. For {ACE}, adding the guidelines in the instruction does not offer a further advantage, where \textbf{NoGuideline} and \textbf{Guideline-PN} achieved a comparable, the best performance, while all other guideline variants do not show to help. On RichERE, however, the benefit of annotation guidelines complements the negative samples', where \textbf{Guideline-PN} and \textbf{Guideline-PS} achieve around 25\% gain on AC over \textbf{NoGuideline}. We notice that RichERE is annotated with a smaller training set but defines more fine-grained event schemas than ACE; for example, the courser-grained \texttt{Transport} event type in ACE is represented by two event types, i.e., \texttt{TransportPerson} and \texttt{TransportArtifact}. As the guideline provides not only a detailed description of an event type but also a comparison with similar ones (Table~\ref{tab:guideline-examples}), the LLM can leverage this information for better EE performance.

\begin{table*}[th!]
\centering
\setlength{\tabcolsep}{4pt}
\renewcommand{\arraystretch}{1.2}
\definecolor{rowgray}{gray}{0.97} 
\definecolor{rowgray2}{gray}{1.0} 
\definecolor{lightblue}{RGB}{173, 216, 250}
\resizebox{\textwidth}{!}{%
\begin{tabular}{lcccc|cccc|cccc|cccc}
\toprule
\multirow{2}{*}{\textbf{Experiments}} & \multicolumn{4}{c|}{\textbf{ACE w/o NS}} & \multicolumn{4}{c|}{\textbf{ACE w/ NS}} & \multicolumn{4}{c|}{\textbf{RichERE w/o NS}} & \multicolumn{4}{c}{\textbf{RichERE w/ NS}} \\
\cmidrule(lr){2-5} \cmidrule(lr){6-9} \cmidrule(lr){10-13} \cmidrule(lr){14-17}
& \textbf{TI} & \textbf{TC} & \textbf{AI} & \textbf{AC} & \textbf{TI} & \textbf{TC} & \textbf{AI} & \textbf{AC} & \textbf{TI} & \textbf{TC} & \textbf{AI} & \textbf{AC} & \textbf{TI} & \textbf{TC} & \textbf{AI} & \textbf{AC} \\
\midrule
\rowcolor{lightblue}\textbf{NoGuideline}      & 10.60  & 10.60  & 5.19  & 3.68  & 31.64  & 31.64  & 25.91  & 24.22  & 19.87  & 19.87  & 13.34  & 11.69  & 36.29  & 36.29  & 28.15  & 25.58  \\
\rowcolor{rowgray2}\textbf{Guideline-H}      & 29.01  & 29.01  & 16.37  & 14.78  & 32.62  & 32.62  & 25.35  & 22.87 & --  & --  & --  & --  & --  & --  & --  & -- \\
\rowcolor{rowgray}\textbf{Guideline-P}      & \underline{36.91}  & \underline{36.91}  & \underline{24.17}  & \underline{21.24}  & \underline{56.99}  & \underline{56.99}  & \textbf{43.44}  & \textbf{40.51}  & \textbf{40.28}  & \textbf{40.28}  & \textbf{21.97}  & \textbf{18.33}  & \underline{62.04}  & \underline{62.04}  & \underline{46.33}  & \underline{42.03}  \\
\rowcolor{rowgray2}\textbf{Guideline-PN}     & 30.94  & 30.94  & 19.27  & 17.64  & \textbf{60.29}  & \textbf{60.29}  & \underline{42.88}  & \underline{39.95}  & \underline{31.23}  & \underline{31.23}  & \underline{19.48}  & \underline{17.51}  & \textbf{67.16}  & \textbf{67.16}  & \textbf{47.85}  & \textbf{43.39}  \\
\rowcolor{rowgray}\textbf{Guideline-PS}     & \textbf{40.53}  & \textbf{40.53}  & \textbf{28.03}  & \textbf{26.12}  & 55.1  & 55.1  & 41.57  & 38.91  & 26.16  & 26.16  & 16.64  & 15.19  & 58.95  & 58.95  & 42.79  & 38.1  \\
\rowcolor{rowgray2} \textbf{Guideline-PN-Int} & 34.11  & 34.11  & 22.73  & 21.18  & 28.31  & 28.31  & 23.82  & 22.37  & 25.73  & 25.73  & 16.75  & 14.6  & 33.59  & 33.59  & 28.06  & 26.0  \\
\rowcolor{rowgray} \textbf{Guideline-PS-Int} & 30.04  & 30.04  & 19.69  & 16.9  & 27.96  & 27.96  & 21.55  & 20.37  & 23.33  & 23.33  & 15.35  & 13.38  & 34.92  & 34.92  & 27.31  & 25.04  \\
\bottomrule
\end{tabular}%
}
\caption{Evaluation results (\%) on full test data, for end-to-end EE tasks, trained on 2000 train data samples.}
\label{tab:train2000_table}
\end{table*}

\begin{table}[t!]

\setlength{\tabcolsep}{4pt}
\renewcommand{\arraystretch}{1.2}
\small
\definecolor{rowgray}{gray}{0.97} 
\definecolor{rowgray2}{gray}{1.0} 
\definecolor{lightblue}{RGB}{173, 216, 250}
\resizebox{\linewidth}{!}{%
\begin{tabular}{lcccc|cccc}
\toprule
 & \multicolumn{4}{c|}{\textbf{ACE w/ NS}} & \multicolumn{4}{c}{\textbf{RichERE w/ NS}} \\
\cmidrule(lr){2-5} \cmidrule(lr){6-9}
& \textbf{TI} & \textbf{TC} & \textbf{AI} & \textbf{AC} & \textbf{TI} & \textbf{TC} & \textbf{AI} & \textbf{AC} \\
\midrule
\rowcolor{lightblue}\textbf{NoGuide}      & \textbf{37.08}  & \textbf{37.08}  & \textbf{21.53}  & \textbf{19.18}  & 24.98  & 24.98  & 15.05  & 13.15  \\
\rowcolor{rowgray2}\textbf{H}      & 29.00  & 29.00  & 17.93  & 16.34  & --  & --  & --  & --  \\
\rowcolor{rowgray}\textbf{P}      & 27.95  & 27.95  & 15.94  & 14.21  & 23.93  & 23.93  & 13.56  & 12.71  \\
\rowcolor{rowgray2}\textbf{PN}     & 29.60  & 29.60  & 17.87  & 15.92  & \underline{27.43}  & \underline{27.43}  & \textbf{17.10}  & \textbf{15.28}  \\
\rowcolor{rowgray}\textbf{PS}     & \underline{29.85}  & \underline{29.85}  & \underline{19.49}  & \underline{17.04}  & 19.61  & 19.61  & 11.77  & 10.48  \\
\rowcolor{rowgray2}\textbf{PN-Int} & 24.34  & 24.34  & 14.08  & 12.56  & \textbf{27.59}  & \textbf{27.59}  & \underline{16.21}  & \underline{14.47}  \\
\rowcolor{rowgray}\textbf{PS-Int} & 22.51  & 22.51  & 13.59  & 12.48  & 18.99  & 18.99  & 10.67  & 9.56  \\
\bottomrule
\end{tabular}%
}
\caption{Evaluation results (\%) for end-to-end EE tasks on full test data, averaged over three runs using 100 training samples. {We did not experiment with the ``w/o NS'' setting because the model performance with 100 training samples is negligible for all variants.}}
\label{tab:train_100_results}
\vspace{-5pt}
\end{table}

\subsection{RQ2: Are machine-generated annotation guidelines effective?}
Interestingly, from Table~\ref{tab:full_train_table}, we noticed that the guidelines provided by the ACE annotators do not yield a performance gain and that the machine-generated guideline variants are not equally effective. Specifically, \textbf{Guideline-H} achieves a comparable performance in \textbf{w/o NS} and an inferior one in \textbf{w/ NS} on ACE; \textbf{Guideline-PN-Int} and \textbf{Guideline-PS-Int} provide either no or limited performance gain in both \textbf{w/o NS} and \textbf{w/ NS} settings, while \textbf{Guideline-P} and \textbf{Guideline-PS} are not consistently better than \textbf{NoGuideline}. \textbf{Guideline-PN} shows to be the most stable, outperforming \textbf{NoGuideline} on RichERE and performing comparably to the best model on ACE.

Qualitatively, as shown in Table~\ref{tab:guideline-examples}, the human-written guidelines (\textbf{Guideline-H}) lack explicit contrasts, making event boundaries ambiguous—for instance, \texttt{Transport} (a movement event) and \texttt{Extradite} (a justice event) both involve relocation, yet the fact that only the latter is legally enforced is not clarified in the guidelines.
\textbf{Guideline-P} provides examples and edge cases of the target event, but these may not be sufficient for the model to distinguish between similar event types. While both \textbf{Guideline-PS} and \textbf{Guideline-PN} have supplied this comparison, \textbf{-PS} shows to be limited by focusing on only sibling differentiations (e.g., \texttt{Extradite} vs. \texttt{Convict}). Finally, surprisingly, the two \textbf{-Int} variants, despite being comprehensive, lead to mixed results. We observed that models tend to overfit to these comprehensive instructions. In contrast, training the models with 5 diverse guidelines per event type as in \textbf{-PN} and \textbf{-PS} avoids this issue, which shares a similar finding as \citet{cai-etal-2024-improving-event, sainz2024gollie}.


\subsection{RQ3: Are the annotation guidelines helpful when there is only a small amount of training data?}
With 2000 samples (Table~\ref{tab:train2000_table}), \textbf{Guideline-P}, \textbf{Guideline-PN} and \textbf{Guideline-P} improve \textbf{NoGuideline} on ACE and RichERE w/o NS by up to 30\% TC and 20\% AC. Unlike our observation on the full-training setting, this trend also holds in \textbf{ACE w/ NS}, where guidelines provide a similar advantage. 
Excitingly, the results also show that annotation guidelines can compensate for limited training data, enabling models trained with only 2000 samples to achieve performance comparable to full-data training. For example, on ACE, \textbf{Guideline-P w/ NS (2k)} outperforms \textbf{NoGuideline w/o NS (full)} by 10\% AC; on RichERE, \textbf{Guideline-PN w/ NS (2k)} outperforms \textbf{NoGuideline (full)} by 18\% AC in ``w/o NS '' and 12\% AC in ``w/ NS''.


However, when training data is reduced to 100 samples (Table~\ref{tab:train_100_results}), the benefits become dataset-dependent. In \textbf{ACE w/ NS}, \textbf{NoGuideline} slightly outperforms guideline-based models, suggesting that with extremely limited data, the model resorts to memorization rather than learning schema constraints. In contrast, in \textbf{RichERE w/ NS}, which has more diverse and fine-grained event structures, guidelines remain beneficial—\textbf{Guideline-PN} surpasses \textbf{NoGuideline} by 2\% AC, indicating that guidelines help in settings where direct memorization is insufficient.

\begin{table*}[t]
\centering
\setlength{\tabcolsep}{3.5pt}
\renewcommand{\arraystretch}{1.2}
\definecolor{rowgray}{gray}{0.97} 
\definecolor{rowgray2}{gray}{1.0} 
\definecolor{lightblue2}{RGB}{173, 216, 240} 
\definecolor{lightblue}{RGB}{173, 216, 250} 
\resizebox{\textwidth}{!}{%
\begin{tabular}{lcccc|cccc|cccc|cccc}
\toprule
\multirow{2}{*}{\textbf{Experiments}} 
& \multicolumn{4}{c|}{\textbf{RichERE w/o NS → ACE}} 
& \multicolumn{4}{c|}{\textbf{RichERE w/ NS → ACE}} 
& \multicolumn{4}{c|}{\textbf{ACE w/o NS → RichERE}} 
& \multicolumn{4}{c}{\textbf{ACE w/ NS → RichERE}} \\
\cmidrule(lr){2-5} \cmidrule(lr){6-9} \cmidrule(lr){10-13} \cmidrule(lr){14-17}
& \textbf{TI} & \textbf{TC} & \textbf{AI} & \textbf{AC} 
& \textbf{TI} & \textbf{TC} & \textbf{AI} & \textbf{AC} 
& \textbf{TI} & \textbf{TC} & \textbf{AI} & \textbf{AC} 
& \textbf{TI} & \textbf{TC} & \textbf{AI} & \textbf{AC} \\
\midrule
\rowcolor{lightblue}\textbf{NoGuideline}      & 29.55  & 29.55  & 21.34  & 16.60  & 44.10  & 44.10  & 33.91  & 25.17  & 33.41  & 33.41  & 24.34  & 22.68  & 37.19  & 37.19  & 27.74  & 25.87 \\
\rowcolor{rowgray2}\textbf{Guideline-P}      & 31.78  & 31.78  & 22.51  & 15.90  & 61.69  & 61.69  & 39.83  & 27.93  & \textbf{42.95}  & \textbf{42.95}  & \textbf{31.61}  & \textbf{27.79}  & 54.72  & 54.72  & 38.63  & 35.00  \\
\rowcolor{rowgray}\textbf{Guideline-PN}     & \textbf{40.12}  & \textbf{40.12}  & \textbf{27.78}  & \textbf{19.77}  & \underline{63.97}  & \underline{63.97}  & \textbf{48.74}  & \textbf{36.24}  & 41.72  & 41.72  & 29.54  & 26.10  & \underline{64.87}  & \underline{64.87}  & \textbf{48.25}  & \textbf{44.51}  \\
\rowcolor{rowgray}\textbf{Guideline-PS}     & 29.28  & 29.28  & 20.13  & 15.38  & \textbf{64.23}  & \textbf{64.23}  & \underline{44.12}  & \underline{32.84}  & \underline{42.33}  & \underline{42.33}  & \underline{29.93}  & \underline{26.73}  & \textbf{65.54}  & \textbf{65.54}  & \underline{45.57}  & \underline{41.68}  \\
\rowcolor{rowgray2}\textbf{Guideline-PN-Int} & 27.00  & 27.00  & 18.91  & 14.66  & 35.35  & 35.35  & 28.07  & 21.82  & 28.65  & 28.65  & 22.13  & 19.87  & 38.60  & 38.60  & 27.46  & 26.02  \\
\rowcolor{rowgray}\textbf{Guideline-PS-Int} & \underline{31.96}  & \underline{31.96}  & \underline{23.60}  & \underline{19.00}  & 51.71  & 51.71  & 39.36  & 31.34  & 34.33  & 34.33  & 26.65  & 24.24  & 36.85  & 36.85  & 27.69  & 26.19  \\
\bottomrule
\rowcolor{lightblue2}\textbf{In-Distribution}      & 39.57  & 39.57  & 31.05  & 29.73  & 84.15  & 84.15  & 64.99  & 61.96  & 35.11  & 35.11  & 27.16  & 25.32  & 42.27  & 42.27  & 32.38  & 31.56  \\
\bottomrule
\end{tabular}%
}

\caption{{Evaluation of models (\%) in cross-schema generalization. \textbf{In-Distribution} represents the NoGuideline performance when trained and tested on the same dataset and the same setting (w/o or w/ NS). We did not experiment with Guideline-H as RichERE does not come with human-annotated guidelines.}
}
\label{tab:RQ4_f1_scores}
\end{table*}

\subsection{RQ4: Do annotation guidelines improve cross-schema generalization?}
In Table~\ref{tab:RQ4_f1_scores}, we evaluate different variants' generalizability to a new schema in EE. Notably, while ACE and RichERE share the same domain, RichERE has a finer schema design.
{In \textbf{RichERE w/o NS → ACE}, performance remains below the in-distribution baseline. 
While \textbf{Guideline-PN} achieves 40\% TC, nearly matching the in-distribution score, its AC drops by nearly 10\%, likely due to RichERE’s expanded argument roles that do not always align well with ACE’s simpler schema. This suggests that fine-to-coarse schema migration is partially feasible but still faces challenges in argument mapping. Contrastive learning helps mitigate some of this gap, as seen in \textbf{Guideline-PS (w/ NS)}, which improves TC to 64\% and AC to 32\%, highlighting the benefits of structured alignment. In contrast, \textbf{ACE → RichERE} generalizes even better, with \textbf{Guideline-PN (w/ NS)} achieving 64\% TC and 44\% AC, surpassing the in-distribution baseline by over 22\% TC and 12\% AC. This suggests that training on ACE, which has well-defined event boundaries, provides a stronger foundation for adapting to RichERE’s more detailed schema. Since RichERE introduces additional argument roles for certain events in ACE, structured guidelines play a key role in preventing role confusion and ensuring more consistent schema adaptation.}

\subsection{Further Analysis}
\label{sec:analyais}
\begin{table}[t!]

\setlength{\tabcolsep}{4pt}
\renewcommand{\arraystretch}{1.2}
\definecolor{rowgray}{gray}{0.97} 
\definecolor{rowgray2}{gray}{1.0} 
\definecolor{lightblue}{RGB}{173, 216, 250}
\definecolor{lightblue2}{RGB}{173, 216, 240} 
\resizebox{\linewidth}{!}{%
\begin{tabular}{lcccc|cccc}
\toprule
 & \multicolumn{4}{c|}{\textbf{ACE}} & \multicolumn{4}{c}{\textbf{RichERE}} \\
\cmidrule(lr){2-5} \cmidrule(lr){6-9}
& \textbf{TI} & \textbf{TC} & \textbf{AI} & \textbf{AC} & \textbf{TI} & \textbf{TC} & \textbf{AI} & \textbf{AC} \\
\midrule
\rowcolor{lightblue}\textbf{NoGuide w/o NS}  & 29.90 & 29.90 & 20.70 & 19.44 & 32.74 & 32.74 & 24.18 & 22.35  \\
\rowcolor{rowgray2}\textbf{PN w/o NS}  & 30.88 & 30.88 & 21.82 & 20.15 & 33.72 & 33.72 & 25.24 & 24.48  \\
\rowcolor{lightblue2}\textbf{NoGuide w/ NS}  & 79.81 & 79.81 & 56.41 & 53.85 & 45.70 & 45.70 & 35.68 & 32.69  \\
\rowcolor{rowgray2}\textbf{PN w/ NS}  & 77.95 & 77.95 & 57.30 & 54.21 & 69.10 & 69.10 & 49.26 & 44.10  \\
\bottomrule
\end{tabular}%
}
\caption{Evaluation results (\%) of LLaMA-3.2-1B-Instruct trained on full ACE and RichERE.}
\vspace{-15pt}
\label{tab:llama_3.2}
\end{table}

\paragraph{Generalization to a Smaller LLM}
We experimented with LLaMA-3.2-1B-Instruct for \textbf{NoGuideline} and the best-performing guideline variant \textbf{Guideline-PN}. 
Results in Table~\ref{tab:llama_3.2} display a consistent observation compared to experiments with the larger LLaMA-3.1-8B model (Table~\ref{tab:full_train_table}). That is,
\textbf{Guideline-PN} achieves a comparable or better result than \textbf{NoGuideline} and shows the advantage of guidelines, particularly on \textbf{RichERE w/ NS}.


\begin{figure}[t!]
    \centering
    \includegraphics[width=0.9\linewidth]{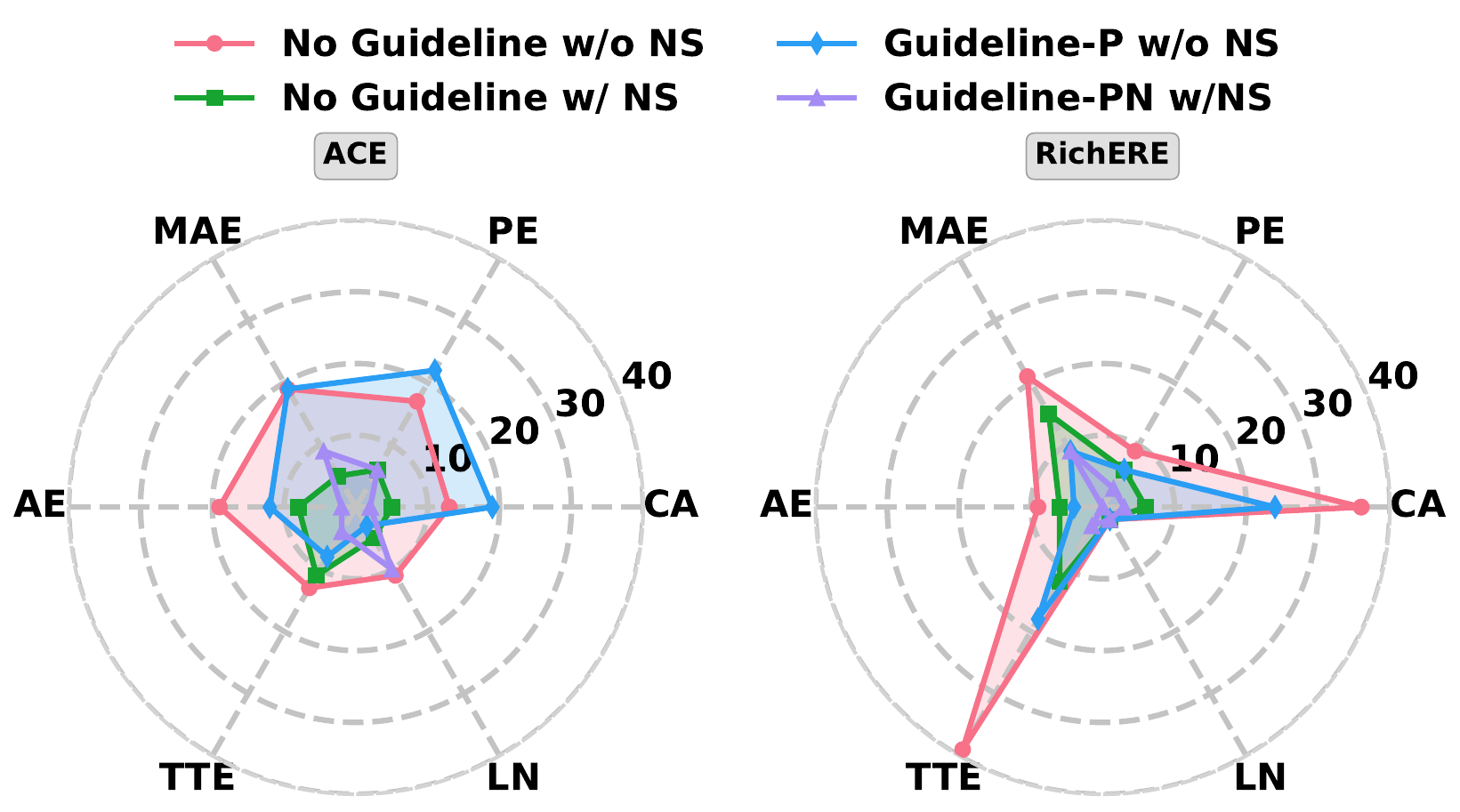}
    \caption{Error categorization: CA (Context Ambiguity), PE (Parsing Errors), MAE (Missing Arguments/Events), AE (Argument Errors), TTE (Type/Trigger Errors), and LN (Label Noise).
    }
    \label{fig:error_cat}
    \vspace{-1.5em}
\end{figure}

\paragraph{Error Analysis}
We randomly selected 100 examples on each dataset where \textbf{NoGuideline w/o NS} made mistakes and compared them with errors made by other variants. The results in Figure~\ref{fig:error_cat} show that, on ACE w/o NS, including the annotation guidelines leads to increasing ungrammatical code outputs and parsing errors (PE), although it dramatically reduces the event type and trigger errors (TTE). In the case of w/ NS, guidelines help in almost all aspects, with the majority of remaining errors being caused by missing arguments or events (MAE) and label noise (LN). On RichERE, however, we observe that for both w/o and w/ NS cases, the annotation guidelines enhance the model performance in all dimensions.

\begin{table}[H]
\centering
\setlength{\tabcolsep}{4pt}
\renewcommand{\arraystretch}{1.2}
\definecolor{rowgray}{gray}{0.97} 
\definecolor{rowgray2}{gray}{1.0} 
\definecolor{lightblue}{RGB}{173, 216, 250}
\definecolor{lightblue2}{RGB}{173, 216, 240}
\resizebox{.8\linewidth}{!}{%
\begin{tabular}{lcccc}
\toprule
\textbf{Setup} & \textbf{TI} & \textbf{TC} & \textbf{AI} & \textbf{AC} \\
\midrule
\rowcolor{lightblue} \textbf{NoGuide w/o NS} & 33.91 & 33.91 & 25.24 & 23.82 \\
\rowcolor{rowgray2} \textbf{PN w/o NS} & 32.38 & 32.38 & 26.19 & 25.32 \\
\rowcolor{lightblue2} \textbf{NoGuide w/ NS} & 46.33 & 46.33 & 36.29 & 33.22 \\
\rowcolor{rowgray2} \textbf{PN w/ NS} & 68.92 & 65.54 & 48.25 & 45.57 \\
\bottomrule
\end{tabular}%
}
\caption{RichERE results with Qwen2.5-Coder-1.5B. Machine-generated guidelines and negative samples consistently improve performance.}
\vspace{-15pt}
\label{tab:qwen_results}
\end{table}

\paragraph{Generalization to a CodeLLM}
{To further assess the robustness and generalizability of our findings beyond the LLaMA family, we conducted additional experiments using \textbf{Qwen2.5-Coder-1.5B-Instruct}, a code-oriented LLM with a distinct architectural design and pretraining objective. We focused specifically on replicating the RichERE experiments (see Table~\ref{tab:llama_3.2}), comparing the \textbf{PN Guideline} variant against the baseline \textbf{NoGuideline}, both with and without negative samples. As shown in Table~\ref{tab:qwen_results}, Qwen2.5-Coder demonstrates substantial improvements. Notably, we observe a 2\% absolute increase in AC performance when guidelines are applied without negative samples (23.82 Vs. 25.32) and a significant 12.35\% gain in TI performance when negative samples are included (46.33 Vs. 68.92). These consistent improvements further reinforce the efficacy of incorporating machine-generated annotation guidelines, highlighting their generalizability across diverse LLM architectures and instruction-tuning paradigms.}
\begin{table}[h!]
\vspace{-7pt}
\centering
\setlength{\tabcolsep}{4pt}
\renewcommand{\arraystretch}{1.2}
\definecolor{rowgray}{gray}{0.97} 
\definecolor{rowgray2}{gray}{1.0} 
\definecolor{lightblue}{RGB}{173, 216, 250}
\definecolor{lightblue2}{RGB}{173, 216, 240} 
\resizebox{.8\linewidth}{!}{%
\begin{tabular}{lcccc}
\toprule
\textbf{Setup} & \textbf{TI} & \textbf{TC} & \textbf{AI} & \textbf{AC} \\
\midrule
\rowcolor{lightblue} \textbf{NoGuide w/o NS} & 31.99 & 30.56 & 22.27 & 21.57 \\
\rowcolor{rowgray2} \textbf{PN w/o NS} & 43.03 & 43.03 & 29.19 & 28.15 \\
\rowcolor{lightblue2} \textbf{NoGuide w/ NS} & 47.95 & 47.08 & 34.01 & 32.50 \\
\rowcolor{rowgray2} \textbf{PN w/ NS} & 62.56 & 62.56 & 42.00 & 40.18 \\
\bottomrule
\end{tabular}%
}
\caption{Speed++ with LLaMA-3.2-1B. Machine-generated guidelines and negative samples consistently improve performance.}

\vspace{-17pt}
\label{tab:speed}
\end{table}

\paragraph{Generalization Beyond the News Domain}
{To study the utility of guidelines for domain generalization, we extended our evaluation to Speed++~\cite{parekh-etal-2024-speed}, a dataset from the epidemic domain containing informal social media posts (tweets), contrasting significantly with the formal newswire text in ACE05 and RichERE. We utilized the same code-format conversion pipeline employed in previous experiments to maintain methodological consistency. For these experiments, we selected the LLaMA-3.2-1B model, aligning with our earlier analysis settings. Although our primary experiments leveraged larger models (e.g., LLaMA-3.1-8B), we opted for the smaller variant due to resource constraints. Nevertheless, this setup provides sufficient validation of our guideline approach across diverse textual domains. Table~\ref{tab:speed} summarizes our findings. Consistent with previous outcomes, we observed substantial improvements when incorporating machine-generated guidelines. Specifically, with negative sampling (w/ NS), TI increased by 14.61\% (47.95 Vs. 62.56), and AC improved by 7.68\% (32.50 Vs. 40.18). These results further underscore the robustness and generalizability of our proposed method, effectively extending its applicability beyond formal news articles to informal social media contexts.}

\paragraph{Effectiveness of Guidelines per Event Type Frequency}
{Figure 3 visualizes the change in AC scores across individual event types for both ACE05 and RichERE. It compares the performance of prompts without guidelines (dashed lines) against those enriched with machine-generated guidelines (solid lines). On average, machine-generated guidelines improve AC scores by +5.47 points on ACE05 (29.73\% Vs. 35.2\%) and +1.86 points on RichERE (25.32\% Vs. 27.18\%).}

{Event types on the y-axis are sorted by training frequency, and the green/red bars show per-event gains or drops. Among the 15 least frequent event types (indices $\ge$15), we observe gains in 8 ACE05 types and 5 RichERE types. For the remaining 5 event types in each dataset show no improvement (+0.0\%) is observed, and a small decline is observed for 2 types in ACE05 and 5 in RichERE.}

{Upon inspection, most of the event types with limited or negative gains have extremely sparse training data, often less than 10 examples (e.g., Marry event), and in some cases as few as 2-3 instances (e.g., Acquit event). In such settings, the model may not receive enough signals to generalize. While our machine-generated guidelines cannot fully compensate for this sparsity, they still improve performance on several rare events, highlighting their potential to support low-resource settings.}

\section{Related Work}
\label{sec:related-work}
\paragraph{LLMs for IE and EE} With the growing capabilities of LLMs, recent efforts have explored their potential in IE~\cite{xu2024large} and studied EE as an auxiliary task. Existing LLM-based IE methods generally fall into two categories: In-Context Learning (ICL) and Supervised Fine-Tuning (SFT). ICL-based approaches~\cite{li-etal-2023-codeie, guo2023retrievalaugmentedcodegenerationuniversal, ashok2023promptnerpromptingnamedentity, wang2023code4struct} rely on providing a few-shot context within prompts, enabling LLMs to infer structured information without explicit parameter updates. 
While being data-efficient, they were found to misinterpret the task specifications~\cite{gao2024eventrlenhancingeventextraction} and suffer from brittle sensitivity to prompt phrasing and example ordering \citep{gao2023exploringfeasibilitychatgptevent}. In addition, they also incur prohibitive costs due to the lengthy reasoning chains especially for complex tasks.
In contrast, SFT-based methods~\cite{lu-etal-2023-pivoine, wang2023instructuiemultitaskinstructiontuning, gui2024instructiebilingualinstructionbasedinformation, zhou2024universalnertargeteddistillationlarge, wei-etal-2024-llms} fine-tune LLMs on annotated datasets, which can significantly improve their EE performance. Our work deepens this line of research and particularly explores the inclusion of annotation guidelines in instructions. While there have been existing works on similar topics, they did not focus on EE \cite{sainz2024gollie} or instruction tuning~\cite{pang-etal-2023-guideline}.

\begin{figure}[t!]
    \centering
    \begin{subfigure}{0.49\linewidth}
        \centering
        \includegraphics[width=\linewidth]{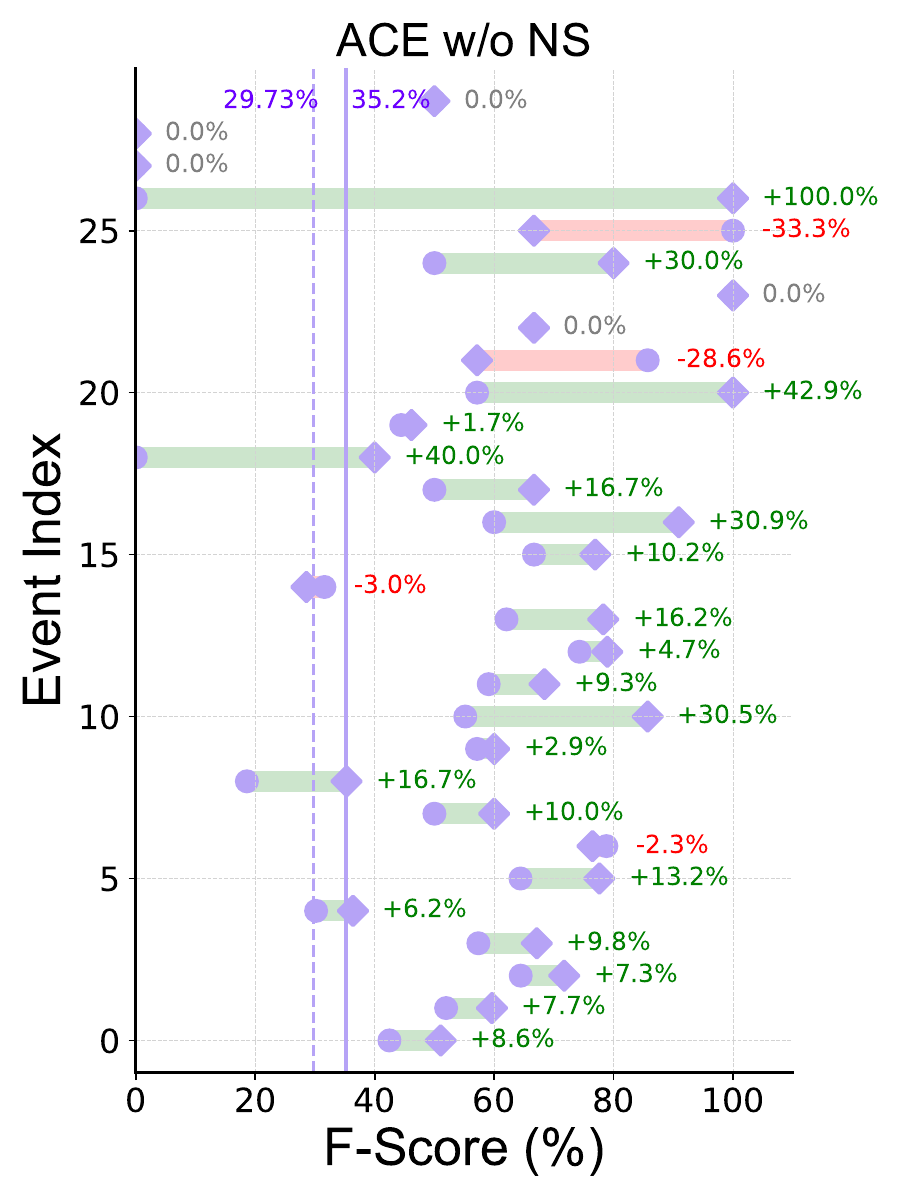}
    \end{subfigure}
    \begin{subfigure}{0.49\linewidth}
        \centering
        \includegraphics[width=\linewidth]{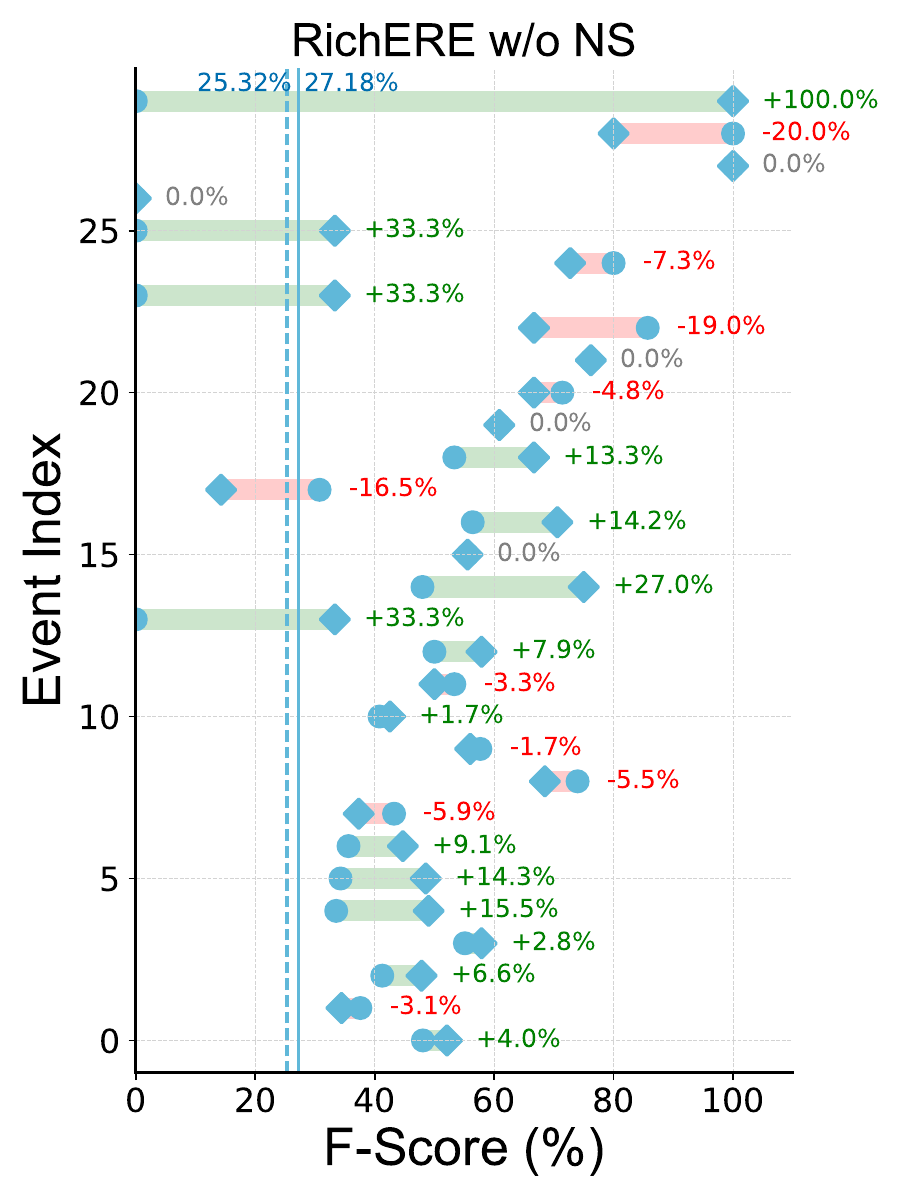}
    \end{subfigure}
    \caption{Impact of guidelines on AC scores per ET, sorted by frequency in the full training set. Smaller index indicate a higher frequency. Green/red bars indicate improvements/declines. Dashed/solid lines denote average AC scores without/with guidelines.
    }
    \label{fig:performance_by_type_frequency}
    \vspace{-12pt}
\end{figure}
\paragraph{Code Prompts for EE} 
While EE tasks are typically represented in texts, code-based prompting has emerged as a promising alternative, leveraging structured representations to enhance schema adherence. Early works have applied code-style prompts to event argument extraction~\cite{wang2023code4struct} and other IE tasks~\cite{li-etal-2023-codeie}, demonstrating potential but often underperforming compared to SFT-based models due to the absence of fine-tuning. EventRL~\cite{gao2024eventrlenhancingeventextraction} utilizes outcome supervision with specific reward functions to reduce information mismatch and hallucination. KnowCoder~\cite{sainz2024gollie, li-etal-2024-knowcoder} addresses this limitation by introducing a comprehensive schema representation in code format, integrating taxonomies, constraints, and structured definitions. 
Complementary to these works, we study generating annotation guidelines to enhance the instruction tuning of LLMs for code-formatted EE and demonstrate their effectiveness.
\section{Conclusion}
We demonstrate that incorporating structured annotation guidelines improves the instruction-tuning of LLMs for EE, bridges the data gap when only a limited amount of training data is available, and enhances the model's cross-schema generalization. Our explorations of guideline generation also highlight the promise of automatically generating effective instructions.

\section{Limitations}
While our study demonstrates the benefits of structured annotation guidelines for event extraction, several limitations remain. 
{
First, while our evaluation includes ACE and RichERE, both within the news domain, and an additional experiment on Speed++ from the epidemic domain, our study still lacks coverage of other domains such as biomedical or legal texts. This limited domain diversity may affect the generalizability of our findings.
}
Future work should assess whether schema differences in other domains exhibit similar trends. Second, while we analyze guideline length and diversity, we do not explicitly optimize guideline generation, leaving open the question of how to best balance conciseness and informativeness. Exploring adaptive methods that retrieve or refine guidelines dynamically during training and inference could further improve efficiency. Lastly, our study primarily focuses on instruction-tuning an LLM with predefined event schemas;     however, real-world applications often require handling previously unseen event types. Investigating how structured guidelines can aid zero-shot or few-shot event extraction remains an important avenue for future research. { Additionally, future work could explore how machine-generated guidelines can assist human annotators during dataset creation, potentially improving annotation consistency and inter-annotator agreement in complex event extraction tasks.}

\section*{Acknowledgements}
This project was sponsored by the College of Computing and Engineering and the Department of Computer Science at George Mason University. This project was also supported by resources provided by the Office of Research Computing at George Mason University (URL: \url{https://orc.gmu.edu}) and funded in part by grants from the National Science Foundation (Award Number 2018631).
\bibliography{custom}


\appendix

\section{Preprocessing and Data Sampling}
\label{sec:appendix_preprocessing}
For both datasets, ACE and RichERE, we follow the TextEE standardization \cite{huang2024textee} and formulate them as sentence-level EE tasks. We use the ``split 1'' data split of TextEE, but only sample a subset of 100 examples from its development (dev) set for better training efficiency. Specifically, we ensure that for each event type, two event instances will be included in our dev set, prioritizing those with larger coverages of arguments, with the remaining being examples with no event occurrences. The datasets are then converted to the code format shown in Figure~\ref{fig:overview}. Table~\ref{tab:data_stats} summarizes dataset statistics.
\begin{table}[th!]
\centering
\definecolor{rowgray}{gray}{0.97} 
\definecolor{rowgray2}{gray}{1.0} 
\resizebox{\columnwidth}{!}{%
\begin{tabular}{>{\centering\arraybackslash}p{3cm}>{\centering\arraybackslash}p{2cm}>{\centering\arraybackslash}p{1.5cm}>{\centering\arraybackslash}p{3cm}}
\toprule
\textbf{Dataset} & \textbf{\#Event Types} & \textbf{\#Role Types} & \textbf{\#Instances\newline (train/dev/test)} \\ 
\midrule
\textbf{ACE05~\cite{doddington-etal-2004-automatic}} & 33 & 22  & 16531/1870/2519  \\
\textbf{RichERE~\cite{song-etal-2015-light}} & 38 & 35  & 9105/973/1163 \\
\bottomrule
\end{tabular}
}
\caption{Dataset statistics. For efficiency purposes, in our experiments, we curated a subset of 100 examples as our development (dev) set.}
\label{tab:data_stats}
\end{table}

To perform the low-data experiments (RQ3), we additionally create the following subsets of the full training set for each dataset. \textbf{Train2k} includes uniformly sampled 2,000 examples from the full training set. \textbf{Train100-1/2/3} are three distinct subsets including 100 examples from the full training set, each of which was selected following the same procedure as how we prepare the dev set, ensuring all event types are included and prioritizing instances covering more arguments.

\section{Selection of Negative Examples for Guideline Generation}
\label{sec:neg_details}
For each event type, we selected 15 negative examples using a fixed random seed to ensure reproducibility. These negative examples were sampled from event instances belonging to types other than the target event type, as required by the definition of Guideline-PN. Each of these examples was used to inform the LLM that a given input does not express the event type in question. In our preliminary explorations, we did not observe any obvious discrepancy when we repeated the guideline generation with different random selections of 15 negative examples.

\section{Evaluation Methodology and Metrics}
\label{sec:app_additional_details}
\paragraph{Evaluation Methodology.} 
Our methodology contrasts with GoLLIE~\cite{sainz2024gollie}, which follows a pipeline-based structure and selectively includes only parent event types in its prompts, limiting granularity in event representation. For argument extraction, GoLLIE further restricts schema inclusion to sibling event types, introducing manual design choices that reduce automation and scalability. To ensure fair and comprehensive evaluation, we adopt a methodology that enumerates all possible event types for each test and development sample during prompt construction. Unlike setups where only the gold-standard event schema is included in the prompt, we avoid implicit event detection bias—if the correct event type were provided, the model would not need to identify the event type itself and could directly extract arguments, which would not reflect its real performance on real-world data. Due to these fundamental differences in methodology, we do not compare our results with GoLLIE.

\section{Prompt Design and Model Training}
\paragraph{Model.} 
We conducted experiments on an instruction-tuned LLaMA-3-8B model, selected for its demonstrated proficiency in processing structured code-based inputs and generating coherent outputs. For parameter-efficient adaptation, we implement RSLoRA \cite{kalajdzievski2023rankstabilizationscalingfactor}, applying LoRA transformations to all linear layers in the transformer blocks following the methodology of \citet{dettmers2024qlora}. Key hyperparameters—including LoRA rank (64), scaling factor $\alpha$ (128), and batch size (32)—were determined through preliminary experiments to balance computational efficiency with model performance. The models were trained for 10 epochs using a single NVIDIA A100 GPU (80GB VRAM), with early stopping triggered after three consecutive validation steps without improvement. We adopt a cosine learning rate scheduler with an initial rate of 1e-5 and a warmup period of 350 steps. Input sequences are padded to 3,000 tokens to maintain consistency while accommodating long-form code structures. To ensure reproducibility and minimize memory fragmentation, we implement deterministic padding and truncation strategies.

\paragraph{Prompt Design.}
\label{sec:prompt-design}
We adopt a structured prompt format consisting of four components: (i) task instruction,(ii) event schema, (iii) input text, and (iv) expected output, formatted as a structured event representation. Our approach follows a schema-first prompting strategy, where event definitions are explicitly encoded in a structured format to enhance model comprehension of event relations and argument constraints. For each input instance, a randomly sampled guideline definition is used to annotate the event schema, ensuring that the model is exposed to multiple rephrasings rather than memorizing and overfitting on a static definition. Formally, we prepare the input sequence as follows: ``\texttt{[BoS] \$-task\_instruction $(I)$ \$-annotated event schema $({E}_e)$~\$-input\_sample~$(X_i)$~[EoS]}'' where  the event schema ${E}_e$ for an event $e$ is annotated with one of the generated guideline definitions. 



\begin{lstlisting}[style=customjson, caption={Prompt example for generating Guideline-P, Guideline-PN, and Guideline-PS.}, label={lst:guidelines1}, aboveskip=10pt, belowskip=10pt]
You are an expert in annotating NLP datasets for event extraction. Your task is to generate "detailed" annotation guidelines for the event type Acquit which is a child event type of super class JusticeEvent.

Input Format will be as following
```
Event Schema:
Event Name and its parent class
Arguments:
Arguments separated by new lines. If there are no arguments None will be given.

Examples
```
Instructions:
1) Identify and list all unique arguments related to the event type.
2) Define the event type and each argument. You can take help of examples below to understand the events and their arguments. 
3) Please remember that the examples may not cover all the arguments in the list. In some cases, you may not have arguments at all, in such cases, you can have an empty list for arguments. 
4) For each definition, provide 5 illustrative definitions in JSON format. For events you can add example triggers and the explanation of the events such as edge cases and other critical details starting with "The event can be triggered by ... ". Similarly for arguments also you can add examples, and detailed information for them including any edge case or domain knowledge starting with "Examples are ... ".
5) Remember to not generate any additional information such as examples, etc. and strictly follow the output format shown below.
6) Remember also to add detailed information for the events and arguments so that the annotators who are not familiar with machine learning and NLP can still solve the task. Remember to add required domain knowledge and please cover the edge cases when possible.
7) Remember that while generating examples for the event or attributes you should generate diverse set of triggers or argument values rather than picking them from the examples I have provided for each of the 5 generated guidelines.

Output Format:
{
  "Event Definition": [
    "Definition 1",
    "Definition 2",
    "Definition 3",
    "Definition 4",
    "Definition 5"
  ],
  "Arguments Definitions": {
    "Argument1": [
      "Definition 1",
      "Definition 2",
      "Definition 3",
      "Definition 4",
      "Definition 5"
    ],
    "Argument2": [
      "Definition 1",
      "Definition 2",
      "Definition 3",
      "Definition 4",
      "Definition 5"
    ]
    // Add additional arguments as necessary
  }
}

Event Schema:
Acquit which is a child event type of super class JusticeEvent
Arguments:
Argument 1 -> adjudicator
Argument 2 -> defendant

Example 1
### Input Text ###
Sentence 1.
### Event Trigger ###
[event trigger]
### Event Arguments ###
For argument "defendant" extracted spans ['x']
For argument "adjudicator" extracted spans ['y']

Example 2
### Input Text ###
Sentence 2.
### Event Trigger ###
[event trigger]
### Event Arguments ###
For argument "defendant" extracted spans ['a']

(...)
\end{lstlisting}

\paragraph{Prompt for Generating Consolidated Guidelines.}The exact prompts used for generating consolidated guidelines - Guideline-PN-Int, and Guideline-PS-Int is shared below

\begin{lstlisting}[style=customjson, caption={Prompt example for generating consolidated guidelines: Guideline-PN-Int, and Guideline-PS-Int.}, label={lst:guidelines2}, aboveskip=10pt, belowskip=10pt]
You are an expert in summarizing NLP event extraction guidelines. Your goal is to consolidate multiple detailed descriptions into a single concise, comprehensive "Intergrated" guideline.

### Input Format ###
Event Type: Event Type Name
```json
{
  "Event Definition": [
    "Definition 1",
    "Definition 2",
    "Definition 3",
    "Definition 4",
    "Definition 5"
  ],
  "Arguments Definitions": {
    "mention": [
      "Definition 1",
      "Definition 2",
      "Definition 3",
      "Definition 4",
      "Definition 5"
    ],
    "Argument1": [
      "Definition 1",
      "Definition 2",
      "Definition 3",
      "Definition 4",
      "Definition 5"
    ],
    // Add additional arguments as necessary
  }
}
```

### Task ###
1. Integrated the 5 definitions under "Event Definition" into a single definition:
   - Highlight all critical points and examples from the five definitions.
   - Ensure the description is concise, comprehensive, and clear, using formal language that non-experts can understand.

2. Do the same for each argument under "Arguments Definitions," producing a single intergrated definition for each. 

### Output Format ###
```json
{
  "Event Definition": "Consolidated intergrated guideline for the event type.",
  "Arguments Definitions": {
    "mention": "Consolidated intergrated guideline for the mention argument.",
    "Argument1": "Consolidated intergrated guideline for Argument1.",
    "Argument2": "Consolidated intergrated guideline for Argument2."
    // Add additional arguments as necessary
  }
}
```

### Guidelines to Summarize ###
Event Type: prompt_Acquit(JusticeEvent)
```json
{
    "Acquit(JusticeEvent)": {
        "description": [
            "Definition 1",
            "Definition 2",
            "Definition 3",
            "Definition 4",
            "Definition 5"
        ]
    },
    "attributes": {
        "mention": "The text span that triggers the event."
        "adjudicator": [
            "Definition 1",
            "Definition 2",
            "Definition 3",
            "Definition 4",
            "Definition 5"
        ],
        "defendant": [
            "Definition 1",
            "Definition 2",
            "Definition 3",
            "Definition 4",
            "Definition 5"
        ]
    }
}
```
\end{lstlisting}

\section{Dataset Examples Across Multiple Guideline Settings}
\label{sec:app_dd}
The below JSON example illustrates an event extraction task from the ACE dataset under the No Guideline setting. It defines how structured events are extracted from text, specifying event triggers, types, arguments, and roles. The instruction explains the task, the input provides a natural language sentence and its conversion into a structured Python-style format. The output presents the extracted event, including its trigger ("extradited") and associated arguments (e.g., "government" as the agent, "him" as the person).

\begin{lstlisting}[style=customjson, caption={Dataset example from ACE-05 with no guidelines.}, label={lst:guidelines3}, aboveskip=10pt, belowskip=10pt]
{
  "doc_id": "APW_ENG_20030306.0191",
  "wnd_id": "APW_ENG_20030306.0191-6",
  "instance_id": "821",
  "dataset_name": "ace05-en",
  "task_type": "E2E",
  "is_auth": "0",
  "instruction": "# This is an event extraction task where the goal is to extract structured events from the text. A structured event contains an event trigger word, an event type, the arguments participating in the event, and their roles in the event. For each different event type, please output the extracted information from the text into python-style dictionaries where the first key will be 'mention' with the value of the event trigger. Next, please output the arguments and their roles following the same format. The event type definitions and their argument roles are defined next.",
  "input": "# The following lines describe the task definition\n\n@dataclass\nclass Extradite(JusticeEvent):\n    mention: str\n    agent: List\n    destination: List\n    origin: List\n    person: List\n\n# This is the text to analyze\ntext = \"The post-Milosevic government later extradited him to the U.N. war crimes tribunal in The Hague, the Netherlands.\"\n\n# The list called result should contain the instances for the following events according to the guidelines above:\nresult = \n",
  "output": "[Extradite(\n    mention=\"extradited\",\n    person=[\"him\"], \n    destination=[\"Hague\"], \n    agent=[\"government\"],\n    origin=[]\n)]"
}
\end{lstlisting}

\paragraph{NoGuideline}
Shown below is an example from the NoGuideline setting in python code format with no doc string and argument definitions.
\vspace{10pt} 
\begin{lstlisting}[style=custompython, aboveskip=-5pt, belowskip=-5pt]
#Task Instruction
# This is an event extraction task where the goal is to extract structured events from the text. A structured event contains an event trigger word, an event type, the arguments participating in the event, and their roles in the event. For each different event type, please output the extracted information from the text into python-style dictionaries where the first key will be 'mention' with the value of the event trigger. Next, please output the arguments and their roles following the same format. The event type definitions and their argument roles are defined next.

#Input
# The following lines describe the task definition

@dataclass
class Extradite(JusticeEvent):
    mention: str
    agent: List
    destination: List
    origin: List
    person: List

# This is the text to analyze
text = "The post-Milosevic government later extradited him to the U.N. war crimes tribunal in The Hague, the Netherlands."

# The list called result should contain the instances for the following events according to the guidelines above:
result = 

#Output
[Extradite(
    mention="extradited",
    person=["him"], 
    destination=["Hague"], 
    agent=["government"],
    origin=[]
)]
\end{lstlisting}

\paragraph{Guideline-PN}
Shown below is an example from the Guideline-PN setting in python code format.
\vspace{10pt} 
\begin{lstlisting}[style=custompython, aboveskip=-5pt, belowskip=-5pt]
#Task Instruction
# This is an event extraction task where the goal is to extract structured events from the text. A structured event contains an event trigger word, an event type, the arguments participating in the event, and their roles in the event. For each different event type, please output the extracted information from the text into python-style dictionaries where the first key will be 'mention' with the value of the event trigger. Next, please output the arguments and their roles following the same format. The event type definitions and their argument roles are defined next.

#Input
# The following lines describe the task definition

@dataclass
class Extradite(JusticeEvent):
    """The event is triggered by the act of transferring a person from one jurisdiction to another for legal proceedings. Example triggers include 'extradite', 'extradition', and 'extraditing'."""
    mention: str  # The text span that triggers the event.
    agent: List  # Examples are 'court', 'government', 'police department'. The agent is the authority or entity responsible for initiating or carrying out the extradition process.
    destination: List  # Examples are 'jurisdiction', 'Hague', 'state'. The destination is the place to which the person is being extradited.
    origin: List  # Examples are 'state', 'headquarters'. The origin is the place from which the person is being extradited.
    person: List  # Examples are 'she', 'him', 'her'. The person is the individual being extradited.

# This is the text to analyze
text = "The post-Milosevic government later extradited him to the U.N. war crimes tribunal in The Hague, the Netherlands."

# The list called result should contain the instances for the following events according to the guidelines above:
result = 

#Output
[Extradite(
    mention="extradited",
    person=["him"], 
    destination=["Hague"], 
    agent=["government"],
    origin=[]
)]
\end{lstlisting}

\paragraph{Guideline-PN-Int}
Similarly, shown below is an example from the Guideline-PN-Int setting in python code format.
\vspace{10pt} 
\begin{lstlisting}[style=custompython, aboveskip=-5pt, belowskip=-5pt]
# The following lines describe the task definition

@dataclass
class Extradite(JusticeEvent):
    """The Extradite event is triggered by the legal process of transferring a person from one jurisdiction to another for legal proceedings, such as facing charges or serving a sentence. This event involves formal actions by legal authorities and the movement of the individual across jurisdictions. Key triggers include terms like 'extradite', 'extradition', and 'extraditing'. It is distinct from events like 'ArrestJail' and 'ReleaseParole', as it specifically involves cross-jurisdictional transfer rather than initial detention or release from custody."""
    mention: str  # The text span that triggers the event.
    agent: List  # The agent is the authority or entity responsible for initiating or carrying out the extradition process, typically a legal or governmental body. Examples include 'court', 'government', and 'police department'.
    destination: List  # The destination is the place to which the person is being extradited, where they will face legal proceedings or serve a sentence. Examples include 'jurisdiction', 'Hague', and 'state'.
    origin: List  # The origin is the place from which the person is being extradited, where they are currently held or from where they are being transferred. Examples include 'state' and 'headquarters'.
    person: List  # The person is the individual being extradited, the subject of the legal transfer. Examples include 'she', 'him', and 'her'.

# This is the text to analyze
text = "The post-Milosevic government later extradited him to the U.N. war crimes tribunal in The Hague, the Netherlands."

# The list called result should contain the instances for the following events according to the guidelines above:
result = 

#Output
[Extradite(
    mention="extradited",
    person=["him"], 
    destination=["Hague"], 
    agent=["government"],
    origin=[]
)]
\end{lstlisting}
\twocolumn

\end{document}